\pdfoutput=1

\documentclass[11pt]{article}

\usepackage[]{latex/acl}
\usepackage{amsmath}

\usepackage{times}
\usepackage{latexsym}

\usepackage[T1]{fontenc}

\usepackage[utf8]{inputenc}

\usepackage{microtype}

\usepackage{inconsolata}

\usepackage{microtype}

\usepackage{amsmath}
\usepackage{graphicx,subcaption}
\usepackage{CJK}
\usepackage{multirow}
\usepackage{booktabs}

\usepackage{algorithm}
\usepackage[noend]{algpseudocode}
\usepackage{enumitem}
\usepackage{placeins}
\usepackage{cancel} 

\title{Unsupervised Text Style Transfer via LLMs and Attention Masking with Multi-way Interactions}

\author{
   Lei Pan$^1$, \textbf{Yunshi Lan$^1$\thanks{*Corresponding author}}, Yang Li$^2$,  Weining Qian$^1$ \\
  $^1$ East China Normal University, $^2$ Alibaba Group\\
   leipan@stu.ecnu.edu.cn\\
        ly200170@alibaba-inc.com, \{yslan,wnqian\}@dase.ecnu.edu.cn 
}

\begin{document}
\maketitle
\begin{abstract}
Unsupervised Text Style Transfer (UTST) has emerged as a critical task within the domain of Natural Language Processing (NLP), aiming to transfer one stylistic aspect of a sentence into another style without changing its semantics, syntax, or other attributes. 
This task is especially challenging given the intrinsic lack of parallel text pairings. 
Among existing methods for UTST tasks, attention masking approach and Large Language Models (LLMs) are deemed as two pioneering methods.
However, they have shortcomings in generating unsmooth sentences and changing the original contents, respectively.
In this paper, we investigate if we can combine these two methods effectively.
We propose four ways of interactions, that are pipeline framework with tuned orders; knowledge distillation from LLMs to attention masking model; in-context learning with constructed parallel examples.
We empirically show these multi-way interactions can improve the baselines in certain perspective of style strength, content preservation and text fluency.
Experiments also demonstrate that simply conducting prompting followed by attention masking-based revision can consistently surpass the other systems, including supervised text style transfer systems.
On Yelp-clean and Amazon-clean datasets, it improves the previously best mean metric by $0.5$ and $3.0$ absolute percentages respectively, and achieves new SOTA results.

\end{abstract}

\section{Introduction}
\label{sec:intro}

Text Style Transfer (TST) is a widely investigated NLP task that aims to transfer one stylistic aspect of a piece of text (e.g., sentiment polarity, formality, politeness, etc.) into another style without changing its semantics, syntax, or other attributes.
Although TST has attracted increased interest from scholars and a number of methods have been developed to solve TST problems~\cite{shang2019semi, zhang2015character, rao2018dear, zhang2020parallel}, these approaches hold an impractical hypothesis that a substantial amount of parallel training instances are well-annotated.
In the absence of a parallel corpus, traditional approaches for supervised learning are inapplicable. 


\label{sec:intro}
\begin{figure*}
    \centering
    \includegraphics[width=\linewidth]{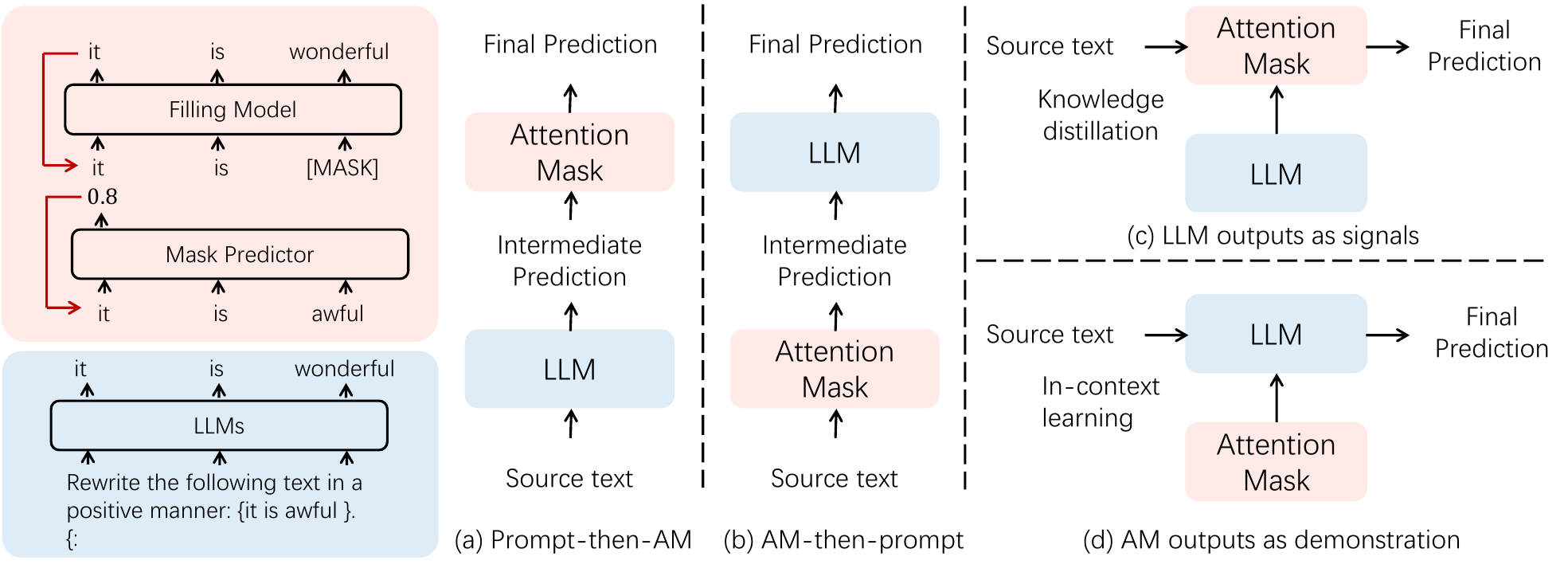}
    \caption{UTST system via LLMs and Attention Masking (AM) with four-way interactions: Prompt-then-AM, AM-then-prompt, knowledge distillation using LLM outputs as signals, and in-context learning using AM outputs as demonstrations.
    The details of attention masking module and LLM-based module are displayed at the left side, where the black arrow denotes propagation and the red arrow denotes back-propagation.}
    \label{fig:methods}
\end{figure*}

Recently, researchers also took efforts to develop a set of systems for Unsupervised Text Style Transfer~\cite{goyal2020multi, lewis-2022-multimodal, luo2023prompt, suzgun2022prompt}.
Among these approaches, we spotlight two types of pioneering methods: attention masking and LLM-based methods.
Attention masking methods utilize attention mechanisms to distinguish the stylistic words that contribute most to the style prediction of a sentence via a fine-tuned classifier and then substitute these identified words to transfer the sentence polarity.
LLM-based methods prompt a general-purpose Large Language Model to transform a sentence into an arbitrary style without additional training or fine-tuning.

However, they both have shortcomings.
The former methods enable modification with controllable edits but the edit to the text is restricted to token-level substitution, which easily generates unnatural and unsmooth expressions.
The latter one enables more flexible text generation but it has a high risk of dramatically changing the content of the original text.
Therefore, an intuitively appealing idea is to combine them together.
But we encounter the core questions: \textit{How to effectively combine attention masking and LLM prompting for UTST tasks? Are there any keys to note when we combine them?}

In this paper, we go deep to explore the possible interactions of LLMs and attention masking, including (1) pipeline framework with tuned orders; (2) knowledge distillation from LLMs to attention masking model; (3) in-context learning with constructed parallel examples.
In particular, we first show the baseline methods to solve UTST tasks.
Then we introduce the multi-way interactions to combine LLMs and attention masking with detailed implementation.
We conduct experiments with three unparalleled datasets and six UTST challenges and conclude that these multi-way interactions can improve the baselines in certain perspectives of style strength, content preservation, and text fluency. 
Especially, the one-way pipeline of prompting followed by attention masking can achieve the SOTA results on UTST tasks, even compared with supervised TST systems.



In summary, the contributions of this study are as follows:

\begin{itemize}
    \item We innovatively take the interactions of LLMs and attention masking method into the spotlight and discuss the efficient way to combine them for UTST tasks, which may benefit general text generation tasks.
    \item We empirically show UTST systems with combining LLMs and attention masking in four ways can improve the baselines on different evaluation metrics like style strength, content preservation, and text fluency.
    \item We achieve the SOTA results of the mean evaluation metric on two commonly-used style transfer datasets, namely Yelp-clean and Amazon-clean, via prompting followed by attention masking-based revision.
\end{itemize}

\section{Related Work}
For text style transfer, most existing studies hypothesize the existence of parallel transferred text. 
In the task of text style transfer with supervision, these parallel corpora are leveraged to learn a model that can convert text from one style to another while preserving the original content~\cite{shen2017style, prabhumoye2018style, fu2018style, li2018delete, luo2019dual, reif2021recipe}. 
These end-to-end approaches typically involve training on pairs of sentences that are semantically equivalent but stylistically distinct, allowing the model to capture the nuances of each style.
However, in real-world scenarios, we frequently encounter cases where the parallel data is unavailable.
To solve such unsupervised text style transfer tasks, there are two lines of mainstream approaches, that aim to implicitly or explicitly model the style-related words in the text respectively and then generate a text in the target style \cite{madaan2020politeness, malmi-etal-2022-text, reid2021lewis, mallinson2022edit5, wang2022text, vincent2008extracting, hu2017toward, kingma2013auto, fu2018style, john2018disentangled, li2020dgst}.

Recent advance in unsupervised text style transfer have leveraged deep learning methods like Variational Autoencoders (VAE) and Denoising Autoencoders (DAE) to modify textual styles while preserving the original content. A notable approach involves learning a latent representation that separates style from content, facilitating the generation of new text in the target style. \citeauthor{hu2017toward} (\citeyear{hu2017toward}) utilize the VAE framework to learn the latent representation of text and employ a style classifier to discern the style attribute vector. Similarly, \citeauthor{fu2018style} (\citeyear{fu2018style}) employ an adversarial network to train a content encoder, with the encoded content vector being transformed by a style-specific decoder. \citeauthor{john2018disentangled} (\citeyear{john2018disentangled}) further refined this approach by using VAEs to segregate style and content representations, with the decoder combining both to generate the desired output.


Another line of stream explicitly modeling the style-related information by identifying the stylistic spans from the text then conducting edits to the identified spans.
Attention masking methods have been explored extensively across various tasks, such as sentiment alteration, gender bias mitigation, and political slant adjustment\cite{li2018delete, sudhakar2019transforming}. These methods typically involve disassociating the original style from the content and then amalgamating the sanitized content with the intended style to synthesize new sentences. A pioneering approach in this domain is the ``Delete, Retrieve, Generate'' framework proposed by \cite{li2018delete}, which has been further enhanced by subsequent studies like \cite{sudhakar2019transforming}. These methodologies excel in retaining substantial portions of the original content while transitioning to the target style. However, they have drawbacks, such as compromised fluency in the generated text and a dependence on the retrieval of analogous content across both source and target styles during the training phase.

Currently, multiple studies attempt to leverage LLMs to solve unsupervised text style transfer \cite{tian2023r, luo2023prompt, suzgun2022prompt, reif:acl2022}.
\citeauthor{suzgun2022prompt}~(\citeyear{suzgun2022prompt}) first elaborately design the instruction for text style transfer, which achieves promising results on USTS tasks under zero or few-shot setting.
\citeauthor{reif:acl2022}~(\citeyear{reif:acl2022}) prompt LLMs to acquire a collection of candidate transferred text in the target style, then a re-ranking process is conducted to further rank this candidate text to produce the final prediction.
However, these methods rely solely on the LLMs, which has a high risk of changing the semantics of the original sentence.
In contrast, our method combines the LLMs with the attention masking method and discusses the multi-way interactions.

\section{Task Definition}

UTST is a NLP task aiming to convert the style of the given text without parallel text pairs.
Specifically, we are given two non-parallel corpus $\mathcal{D}^{s_x} = \{X_i\}$ and $\mathcal{D}^{s_y} = \{ Y_j\}$, with pre-defined styles $s_x$ and $s_y$ respectively.
The key of text style transfer is to train a model that enables to transfer a text from $X$ in \textit{source} style $s_x$ to $Y$ in \textit{target} style $s_y$ with preserving the original content.

\section{Baseline UTST System}
\label{sec:baseline}

We highlight two cutting-edge baseline methods for UTST systems, namely \textit{Attention masking method}~\cite{wang:emnlp2022} and \textit{LLM-based method}~\cite{reif:acl2022}.
The former conducts style transfer with attention masking and styled filling, where the attention is derived from a classifier trained on unparalleled data.
The latter prompts LLMs to transfer the text in source style to target style, which relies on the highly adaptive behaviour of LLMs.

\subsection{Attention Masking Method}
\label{sec:attention_mask_method}

Attention masking method consists of a mask predictor and a styled filling model, which take charge of identifying the positions of stylistic words in a text and predicting new stylistic words at the masked positions, respectively.
We display the method in Figure~\ref{fig:methods}.

\noindent \textbf{Mask predictor.} A RoBERTa-based mask predictor takes a source text $X = \{w_1, w_2, ..., w_n \}$ as the input and produces the text with masks.
Due to the absence of parallel data, we train a classifier using $\mathcal{D}^{s_x}$ and $\mathcal{D}^{s_y}$ to judge the style of the text, where the objective is designed as:
\begin{align}
    \mathcal{L}  = - ( & \sum_{X \sim \mathcal{D}^{s_x}} s_x \cdot \log P_{\text{MaskPredictor}}(X)  \nonumber \\
    &+ \sum_{Y \sim \mathcal{D}^{s_y}} s_y \cdot \log P_{\text{MaskPredictor}} (Y)) \label{eq:mask1}.
\end{align}
Regarding inference, following existing UTST studies~\cite{lewis-2022-multimodal}, we utilize attention score in the classifier as the style feature to decide the mask position, where the scaled attention score is higher than a threshold $\alpha$.
After applying the mask vector to $X$, we obtain $X'$.
For simplicity, we denote the above procedure as:
\begin{align}
    \hat{X}^{mask} = \text{MaskPredictor}(X). 
    \label{eq:mask2}
\end{align}

\noindent \textbf{Filling model.} 
A BART-based model fine-tuned on processed $\mathcal{D}^{s_y}$ is applied to recover the masked words to the target style.
To prepare the data for fine-tuning, we mask out some words in $Y$, the attention score of which is larger than $\alpha$ and we treat $Y$ as the target for fine-tuning.
We denote this procedure as:
\begin{align}
    \hat{Y} = \text{FillingModel}(\hat{X}^{mask}). \label{eq:mask3}
\end{align}

\subsection{LLM-based Method}
\label{subsec:llm}

We leverage LLMs to perform text style transfer.
Following existing study~\cite{reif:acl2022}, we frame style transfer as a sentence rewriting task and prompts with only a natural language instruction.
\begin{itemize}[label={}, labelsep=0pt, leftmargin=10pt]
    \item \textbf{Instruction:} \texttt{Rewrite the following text in a [$s_y$] manner: [$X$]}. 
\end{itemize}
Where \texttt{[$s_y$]} and \texttt{[$X$]} are filled with their instantiation.
With the instruction, no parallel data is required for prompting.
We denote this procedure as:
\textbf{\begin{align*}
    \hat{Y} = \text{LLM}(X)
\end{align*}}

\section{Pipeline UTST System}
\label{sec:pipline}

The motivation behind is that LLMs have shown outstanding capability of arbitrary text transfer even under few-shot setting.
But it has the disadvantage of producing less controllability in the properties of the style-transferred text than models trained on the task-specific training data~\cite{reif:acl2022}.
Therefore, we propose intuitive pipeline to edit text by attention mask models and LLMs successively, which could not only leverage LLMs to conduct arbitrary text transfer at absence of parallel text pairs, but also ensure controlability to the semantics of the text by trained models.

\subsection{Prompt-then-AM}
\label{sec:prompt-mask}

We first conduct text style transfer by prompting the LLMs as illustrated in Section~\ref{subsec:llm}, which results in a sentence with the target style.
This would serve as an intermediate prediction, which might encounter over-transfer of text.
Hence, we apply attention masking, which are pre-trained on the non-parallel text pairs, to the intermediate prediction.
To preserve the content of the intermediate prediction, we tune the threshold $\alpha$ to keep more words in the sentences and request the attention masking model to predict masked tokens.

Take the sentence ``\textit{It is awful.}'' as an example, after prompting the LLMs, the input has been transferred into ``\textit{It is unpleasant.}'' as the intermediate prediction.
After further applying the attention masking model to the intermediate prediction, the sentence is further rewritten as ``\textit{It is wonderful.}'' as the final prediction.

\subsection{AM-then-prompt}
\label{sec:mask-prompt}

Alternatively, we first apply attention masking model to edit the original text, which is inherently reliable at producing text that looks like the training corpus.
This serves as an intermediate prediction.
Then we prompt LLMs to rewrite the intermediate prediction with keeping the semantics unchanged, the prompt of which is displayed as follows:
\begin{itemize}[label={}, labelsep=0pt, leftmargin=10pt]
    \item \textbf{Instruction:} \texttt{Refine the following text without changing its semantic: [$X$]}. 
\end{itemize}

Then the outputs of LLMs are extracted as the final prediction.
We still take the sentence ``\textit{It is awful.}'' as an example, the attention masking model transfers the text to ``\textit{It is good.}'' then we request the LLMs to paraphrase the sentence and obtain the final prediction ``\textit{It is wonderful.}''.

\section{Using LLM Outputs as Signals}
\label{sec:llm-signal}

To further fuse the knowledge from LLMs with trained model, we propose to conduct knowledge distillation for text style transfer.
We utilize a LLM as a teacher to generate teaching data and improve the performance of a smaller student model, that is attention masking model, by the generated teaching data.
Instead of directly considering the outputs of LLMs as prediction, distilling knowledge from the LLMs as training signals, which can be considered as a type of label smoothing regularization~\cite{hu:arxiv2022}, makes the student model learns less about the noisy edits to the text.

To this end, we first sample data from $\mathcal{D}^{s_x}$.
Then, we apply LLMs to produce the annotation to the sampled text, which forms a set of parallel text pairs $\mathcal{D}^{par}=\{X^{s_x}_i \rightarrow X^{s_y}_i\}$.
Given a pair of text $\{X^{s_x}_i, X^{s_y}_i\}$ as pseudo annotations, we first employ a modified Levenshtein algorithm\footnote{\url{https://github.com/chrisjbryant/errant}} to identify the minimum replacement edits from the source text to the target text.
Then we convert the edits as the annotated masks for the source text.
This results in the supervision signals for mask predictor, and we denote it as $X^{mask}_i$.
Take the sentence in Figure~\ref{fig:methods} as an example, comparing the original sentence ``\textit{It is awful}'' and the transferred sentence ``\textit{It is wonderful}'' produced via LLM, we annotate ``\textit{awful}'' with a mask.
Hence the supervision signal is ``[0, 0, 1]'', where ``0'' indicates the current token is not a stylistic token or is already a stylistic token in $s_y$, and ``1'' indicates the current token should be masked and replaced as a stylistic token in $s_x$.

In this way, we can fine-tune mask predictor with supervision signals $\mathcal{D}^{1}_{par}=\{X^{s_x}_i \rightarrow X^{mask}_i\}$ where the objective of Equation~(\ref{eq:mask1}) becomes:
\begin{align}
    \mathcal{L}  = -  \sum_{X \sim \mathcal{D}^{1}_{par}} \sum_{w_k \sim X; w_k \text{is masked}}  \log P_{\text{MaskPredictor}}(w_k) \label{eq:mask4}.
\end{align}
Instead of training a classifier to identify the style of $X$ and utilizing the intermediate features to decide the masked token as described in Section~\ref{sec:attention_mask_method}, this method provides immediate supervision signals to train the mask predictor.
Regarding inference, we directly use the predicted masking probability of each token as the output $\hat{X}^{mask}$.

The filling model follows the original implementation but we enhance the fine-tuning by augmenting data $\mathcal{D}^2_{par} = \{X_i^{mask} \rightarrow X^{s_y}_i\}$, where the masks are derived from the pseudo annotation of $X$ rather than $Y$.
This filling model will be utilized to recover the masked words to the target style and we extract $\hat{Y}$ as the transferred sentence.
Comparing with the original filling model, this procedure strengthens it by involving the annotations from LLMs.

\section{Using AM Outputs as Demonstrations}
\label{sec:mask-demo}

To inject the knowledge of $\mathcal{D}^{s_x}$ as well as $\mathcal{D}^{s_y}$ to LLMs and guide it to produce more stylistic sentences as shown in the corpus. 
In-Context Learning (ICL), as a significant prompting strategy, effectively incorporates a small number of demonstrations as part of the input~\cite{ruis2022large} to help LLMs comprehend the tasks.

Multiple studies~\cite{su-2022:Selective,rubin-etal-2022:learning} indicate that a good demonstration should share similarity to the current query in the perspectives of semantic pattern.
Hence, we first encode the current query as well as the text in $\mathcal{D}^{s_x}$ via bge-base-en-v1.5 \cite{xiao2023c}, which is a commonly used sentence-level encoder.
Then, we extract the final hidden state as the vectorized representations and compute the cosine similarity between the current query and sentences in corpus.
We denote the above procedure as:
\begin{align*}
    s_i = \text{CosSim}_{X_i \in \mathcal{D}^{s_x}}(X, X_i),
\end{align*}
where $s_i$ denotes a score of text performing as a demonstration.
We select the sentences with the top-$k$ similarity scores and apply fine-tuned attention masking model to obtain the transferred sentences, which are denoted as $\mathcal{D}^{par}=\{X^{s_x} \rightarrow X^{s_y}\}$.
Next, we prepend the demonstrations to the prompt, which is shown as follows:
\begin{itemize}[label={}, labelsep=0pt, leftmargin=10pt]
    \item \textbf{Instruction:} \texttt{``[$s_x$] Text'': [$X^{s_x}_1$].}  \texttt{``[$s_y]$ Text'': [$X^{s_y}_1$]. } \texttt{...} \texttt{``[$s_x$] Text'': [$X^{s_x}_k$].} \texttt{``[$s_y]$ Text'': [$X^{s_y}_k$]. }
    \item \texttt{Please rewrite the following text into a [$s_y$] sentiment. ``[$s_x$] Text'': [$X$].} \texttt{``[$s_y$] Text'':}
\end{itemize}

As a result, the outputs of LLMs are extracted as the prediction, which is featured with the style of text corpus from $\mathcal{D}^{s_y}$.
\section{Experimental Setup}
\label{sec:experiment}

\subsection{Dataset}

Following existing UTST studies~\cite{reid2021lewis, suzgun2022prompt}, we choose three datasets for our experiments.

\begin{itemize}[leftmargin=*]
 \item \textbf{Yelp}\footnote{\url{https://github.com/shentianxiao/language-style-transfer/blob/master/data/yelp/sentiment.dev.0}} \cite{shen2017style} is a commonly used sentiment polarity classification dataset consisting of review data for Yelp.
 It has $270$K positive and $180$K negative sentences as non-parallel corpus.
 \item \textbf{Amazon}\footnote{\url{https://github.com/lijuncen/Sentiment-and-Style-Transfer}} \cite{li2018delete} is a sentiment classification dataset consisting of Amazon reviews. 
 It comprises $277$K positive and $180$K negative sentences as non-parallel corpus.
 \item \textbf{Politeness}\footnote{\url{https://github.com/tag-and-generate/politeness-dataset}} \cite{madaan2020politeness} is a dataset designed for classifying text as impolite or polite.
 It is derived from the Enron Email corpus and has $200$K non-parallel corpus with either polite or impolite styles.
 It further contains $1$K test sentences.
\end{itemize}




We employ Yelp-clean and Amazon-clean from existing study\footnote{\url{https://github.com/suzgunmirac/prompt-and-rerank/tree/main/datasets}}~\cite{suzgun-etal-2022-prompt} as test data, which are the pre-processed data from the original text.
They both contain $500$ sentences, which are evenly split between positive and negative styles.

\subsection{Comparable Methods}

We take attention masking (\textbf{AM}) method and \textbf{LLM-based} method as baselines.
And we further explore the effect of the multi-way interactions of LLMs and attention masking as described in Section~\ref{sec:pipline}, \ref{sec:llm-signal}, and \ref{sec:mask-demo}, namely \textbf{Prompt-then-AM}, \textbf{AM-then-prompt}, \textbf{LLM-as-signal}, and \textbf{AM-as-demo}.

\subsection{Evaluation metrics}


We follow the prior studies~\cite{post2018call} and measure the UTST methods with the following evaluation metrics:
We employ a RoBERTa model to independently train binary style classifiers on the Yelp, Amazon, and Politeness datasets respectively. 
And we leverage the predicted polarity probability to judge if the generated sentence matches the desired styles, namely accuracy (\textbf{ACC}).
We also employ reference-BLEU (\textbf{r-sBLEU}) to measure the overlap between the generated text and the references if there is any, and self-BLEU (\textbf{s-sBLEU}) to measure the extent to which the system merely replicates the source text.
We further include GPT-2 to measure the smoothness and naturalness of the generated text by computing \textbf{PPL}.

Following \cite{narasimhan2023text, wang2022text}, we also propose a comprehensive \textbf{Mean} metric, combining ACC, s-sBLEU, and a scaled PPL through a geometric mean. 
When calculating the Mean score, the PPL score was exponentially scaled to align its lower-is-better nature with the higher-is-better orientation of the other metrics, normalizing its range to $[0, 100]$\footnote{The scaled PPL score is calculated by $100 * e^{(-0.015 * PPL)}$.}.

\subsection{Implementation Details}

We fine-tuned the pre-trained RoBERTa \cite{liu2019roberta} and BART \cite{lewis2019bart} models in Equation~(\ref{eq:mask2}) and (\ref{eq:mask3}) with specific configurations tailored to their respective tasks and datasets. 
The $\alpha$ in Equation~\ref{eq:mask4} for masking is set as $0.68$.
For RoBERTa, we set the batch size to $64$ and trained the model over $20$ epochs using the AdamW optimizer with a learning rate of $1e-5$, aiming for a balance between computational efficiency and learning effectiveness. In contrast, the BART model's training was adapted to the dataset characteristics: we employed a smaller batch size of $16$, running $10$ epochs for Yelp and Amazon datasets to adequately capture consumer sentiment nuances, and reduced the training to $5$ epochs for the Politeness dataset, which allowed for quicker convergence given its distinct content.
We incorporate \texttt{ChatGLM2-6B}~\cite{du2022glm} as the backbone of the LLMs used in our methods, due to the wide usage and outstanding performance in style transferring~\cite{xuanfan2023systematic, tao2024cat}.
More details can be found in Appendix~\ref{ap:implementation}.


\section{Results and Analysis}
\label{sec:result}

\subsection{Comparison of Different Interactions}

The experimental results presented in Table~\ref{tab:main} provide a comprehensive overview of the performance of various methods for text style transfer across multiple datasets.
We present a summary of our key findings:

(1) For UTST tasks, in comparison, LLMs are more likely to produce more fluent transferred sentences and AM methods have advantages in transferring sentences to targeted style.
Therefore, it is intuitive to combine these two paradigms together.

(2) Considering the different interactions between AM and LLMs, there is no absolute agreement on the different interactions of LLMs and AM methods regarding the performance.
But on five out of six settings, Prompt-then-AM method outperforms the other methods by mean metric.
This indicates that applying prompting followed by AM could effectively balance the fluency and style strength.

(3) AM-as-demo method exhibits the lowest PPL in major settings.
And it improves ACC of LLM-based baseline with observable margin, suggesting that by adding valid demonstrations could help LLMs learn better about the target styles though the upper bound heavily relies on the capability of the LLM.

(4) LLM-as-signal has more advantage in preserving contents of original text than AM method.
It achieves the best \textit{r}-sBLEU and \textit{s}-sBLEU in majority of settings but usually shows poor ACC.
This may because the signals generated via LLMs are initially have poor style strength, distillation from such signals affects the attention masking in a negative way.

\begin{table}[t!]
\centering
\resizebox{\linewidth}{!}{
\begin{tabular}{c | l | c c c c c}
\toprule
\textbf{Dataset} & \textbf{Method}  & \textbf{ACC} ($\uparrow$)& \textbf{\textit{r}-sBLEU} ($\uparrow$) & \textbf{\textit{s}-sBLEU} ($\uparrow$) & \textbf{PPL} ($\downarrow$) & \textbf{Mean} ($\uparrow$)\\ 
\midrule
A{\small MAZON} & LLM-based  & $29$ & $\underline{25.3}$& $\underline{34.0}$& $\underline{74}$& $32.0$\\
\textit{N}$\rightarrow$ \textit{P} & AM & $80$ & $23.0$ & $31.2$& $178$ & $39.4$\\
& Prompt-then-AM & $\mathbf{87}$ & $17.0$ & $22.2$ & $89$ & $\mathbf{45.2}$\\
& AM-then-prompt & $\underline{83}$& $10.0$ & $13.5$ & $78$ & $42.5$\\
& LLM-as-signal & $27$ & $\mathbf{44.2}$ & $\mathbf{70.9}$ & $187$ & $34.7$\\
& AM-as-demo & $56$ & $19.2$ & $20.5$ & $\mathbf{54}$ & $40.3$\\
\midrule 
A{\small MAZON} & LLM-based & $75$& $31.4$ & $34.7$ & $\underline{62}$& $49.7$\\
\textit{P}$\rightarrow$ \textit{N} & AM & $81$& $\underline{35.3}$& $\underline{47.2}$& $123$ & $48.0$\\
& Prompt-then-AM & $\mathbf{97} $& $29.8$ & $32.7$ & $67$ & $\mathbf{55.4}$\\
& AM-then-prompt & $75$ & $15.3$ & $17.5$ & $\mathbf{56}$ & $45.2$\\
& LLM-as-signal & $55$ & $\mathbf{46.7}$ & $\mathbf{74.9}$ & $127$ & $48.3$\\
& AM-as-demo & $\underline{85}$& $31.3$ & $36.1$ & $84$& $49.8$\\
\midrule 
Y{\small ELP} & LLM-based & $67$ & $27.7$ & $33.1$ & $\mathbf{39}$& 51.9\\
\textit{N}$\rightarrow$ \textit{P} & AM & $80$ & $\underline{38.7}$& $\underline{60.4}$& $128$ & 51.7\\
& Prompt-then-AM & $\mathbf{93}$& 26.7& 31.9& 59& $\mathbf{55.4}$\\
& AM-then-prompt & $\underline{84}$& $30.1$ & $36.4$ & $\underline{53}$& 55.2\\
& LLM-as-signal & $31$ & $\mathbf{42.6}$ & $\mathbf{64.2}$ & $119$ & 37.3\\
& AM-as-demo & 81 & 32.5 & 38.8 & 68 & 52.0\\
\midrule 
Y{\small ELP} & LLM-based & $80$& $31.7$ & $37.6$ & $66$ & 51.6\\
\textit{P}$\rightarrow$ \textit{N} & AM & $78$ & $\underline{37.4}$& $\underline{57.0}$& $108$ & 51.6\\
& Prompt-then-AM & $\mathbf{97}$ & $31.2$ & $36.8$ & $\underline{65}$& $\mathbf{57.2}$\\
& AM-then-prompt & $74$ & $19.8$ & $23.3$ & $\mathbf{51}$ & 47.9\\
& LLM-as-signal & $42$ & $\mathbf{43.6}$ & $\mathbf{86.0}$ & $109$ & 49.2\\
& AM-as-demo & $\underline{94}$& $12.5$ & $15.3$ & $\mathbf{51}$ & 51.9\\
\midrule 
Politeness & LLM-based & $66$ & $-$ & $46.7$ & $61$ & 50.9\\
\textit{I}$\rightarrow$ \textit{P} & AM & $\underline{91}$& $-$ & $\mathbf{63.2}$& $181$& 53.6\\
& Prompt-then-AM & $\mathbf{93}$ & $-$ & $42.4$ & $80$ & 55.2\\
& AM-then-prompt & $80$ & $-$ & $44.5$ & $\underline{57}$& $\mathbf{55.7}$\\
& LLM-as-signal & $54$& $-$ & $\underline{55.6}$& $183$ & 38.7\\
& AM-as-demo & $78$& $-$ &$ 40.7$& $\mathbf{51}$& 55.1\\
\midrule 
Politeness & LLM-based & $53$ &  $-$ & $46.2$ &  61& 46.4\\
\textit{P}$\rightarrow$ \textit{I} & AM & $\underline{91}$& $-$ & $\mathbf{66.9}$& $171$  & 55.2\\
& Prompt-then-AM & $\mathbf{95}$& $-$ & $44.2$ &  $74$ & $\mathbf{57.4}$\\
& AM-then-prompt & $74$ & $-$ & $48.2$ & $\mathbf{55}$& 55.3\\
& LLM-as-signal & $47$& $-$ & $\underline{63.2}$& $176$ & 39.1\\
& AM-as-demo & $67$& $-$ & $34.2$& $\underline{60}$& 47.3\\
\bottomrule
\end{tabular}
}
\caption{Results of text style transfer with multi-level interactions on several UTST datasets. 
``\textit{P}$\rightarrow$ \textit{N}'' and ``\textit{N}$\rightarrow$ \textit{P}'' denotes positive style to negative style and negative style  to positive style, respectively.
``\textit{P}$\rightarrow$ \textit{I}'' and ``\textit{I}$\rightarrow$ \textit{P}'' denotes polite style to impolite style and impolite style to polite style, respectively.}
\label{tab:main}
\end{table}

\subsection{Comparison with Other TST Systems}

We further compare our methods based on LLMs and attention masking with other text style transfer systems with supervision or without supervision.
Table~\ref{tab:yelp-clean} displays the results on Yelp-clean.
As we can see, even though under supervised text style transfer setting, a system can be trained using the parallel corpus, it is still short in producing fluent transferred sentences, which results in a relatively low PPL value.
Unsupervised methods with LLMs can achieve impressive results even without any training or fine-tuning on the parallel corpus, but it shows flaws in preserving the original semantics.
Among all the UTST systems, Prompt-then-AM surpasses the other systems on ACC and obtains the highest results of mean metric.




We have the similar observation based on Table~\ref{tab:amazon-clean}, which includes the results on Amazon-clean.
The supervised TST systems have relatively high style strength but low fluency.
The unsupervised TST systems collaborating with diverse LLMs usually generate fluent target sentences but are short in target style.
Overall, our method of Prompt-then-AM can achieve the best mean score with highest style strength.



\begin{table}[t!]
\centering
\resizebox{\linewidth}{!}{
\begin{tabular}{l | c c c c c }
\toprule
\textbf{Method}  & \textbf{ACC} ($\uparrow$)& \textbf{\textit{r}-sBLEU} ($\uparrow$) & \textbf{\textit{s}-sBLEU} ($\uparrow$) & \textbf{PPL} ($\downarrow$) & \textbf{Mean} ($\uparrow$) \\ 
\midrule
\multicolumn{6}{c}{\textit{Supervised Text Style Transfer}} \\
\midrule
$[1]$ CrossAlignment & 73 & 7.8 & 18.3 & 217 & 31.7\\
$[2]$ BackTrans & 95 & 2.0 & 46.5 & 158 & 50.3\\
$[3]$ MultiDecoder & 46 & 13.0 & 39.4 & 373 & 28.6\\
$[4]$ DeleteOnly & 85 & 13.4 & 33.9 & 182 & 41.8\\
$[4]$ DeleteAndRetrieve & 90 & 14.7 & 36.4 & 180 & 44.4\\
$[5]$ UnpairedRL & 49 & 16.8 & 45.7 & 385 & 31.7\\
$[6]$ DualRL & 88 & 25.9 & 58.9 & 133 & 53.5\\
$[7]$ ST (Multi-Class) & 86 & 26.4 & 63.0 & 175 & 52.0\\
$[7]$ ST (Conditional) & 93 & 22.9 & 52.8 & 223 & 49.8\\
$[8]$ B-GST & 81 & 21.6 & 46.5 & 158 & 45.6\\
\midrule
\multicolumn{6}{c}{\textit{Unsupervised Text Style Transfer}} \\
\midrule
$[9]$ Prompt-and-Rerank (GPT2) & 87 & 14.8 & 28.7 & 65 & 51.1\\
$[9]$ Prompt-and-Rerank (GPT-J) & 87 & 23.0 & 47.7 & 80 & 54.9\\
\midrule
Prompt-then-AM & 93& 26.7& 31.9& 59& $\mathbf{55.4}$\\
AM-then-prompt & $84$& $30.1$ & $36.4$ & $53$ & 55.2\\
AM-as-demo & 81 & 32.5 & 38.8 & 68 & 52.0\\
LLM-as-signal & 31 & 42.6 & 64.2 & 119 & 37.3\\
\bottomrule
\end{tabular}
}
\caption{A comparison of existing methods on Yelp-clean (\textit{N}$\rightarrow$ \textit{P}). References: [1] \cite{shen2017style}, [2] \cite{prabhumoye2018style} , [3] \cite{fu2018style}, [4] \cite{li2018delete}, [5] \cite{xu2018unpaired}, [6] \cite{luo2019dual}, [7] \cite{dai2019style}, [8] \cite{sudhakar2019transforming} [9] \cite{reif2021recipe}.
The results of other systems are copied from prior study~\cite{suzgun2022prompt}.
}
\label{tab:yelp-clean}
\end{table}

\begin{table}[]
\centering
\resizebox{\linewidth}{!}{
\begin{tabular}{l  |c c c c c }
\toprule
\textbf{Method}  & \textbf{ACC} ($\uparrow$)& \textbf{\textit{r}-sBLEU} ($\uparrow$) & \textbf{\textit{s}-sBLEU} ($\uparrow$) & \textbf{PPL} ($\downarrow$) & \textbf{Mean} ($\uparrow$) \\ 
\midrule
\multicolumn{6}{c}{\textit{Supervised Text Style Transfer}} \\
\midrule
$[1]$ Style-Embedding& 47& 13.1& 29.0& 287& 25.8\\
$[1]$ CrossAligned& 74& 1.7& 2.4& 96& 33.4\\
$[1]$ DeleteAndRetrieve& 51& 26.7& 53.5& 113& 41.0\\
$[1]$ TemplateBased& 56& 31.0& 65.7& 200& 42.2\\
\midrule
\multicolumn{6}{c}{\textit{Unsupervised Text Style Transfer}} \\
\midrule
$[2]$ GPT-2-Small& 18& 17.7& 38.1& 48& 34.9\\
$[2]$ GPT-2-Medium& 32& 20.1& 38.0& 57& 37.5\\
$[2]$ GPT-2-Large& 28 & 26.0& 51.2& 55& 41.0\\
$[2]$ GPT-2-XL& 32& 22.3& 41.4& 70& 36.1\\
$[2]$ GPT-Neo-1.3B& 31& 10.9& 20.5& 35& 36.9\\
 $[2]$ GPT-Neo-2.7B& 28& 23.7& 45.9& 57&38.8\\
 $[2]$ GPT-J-6B& 33& 27.1& 47.7& 72&38.2\\
\midrule
Prompt-then-AM & 87& 17.0& 22.2& 89& $\mathbf{45.2}$ \\
AM-then-prompt  & 83& 10.0& 13.5& 78& 42.5\\
AM-as-demo  & 56& 19.2& 20.5& 54& 40.3\\
LLM-as-signal & 27& 44.2& 70.9& 187& 34.7\\
\bottomrule
\end{tabular}
}
\caption{A comparison of existing methods on A{\small mazon}-clean (\textit{N}$\rightarrow$ \textit{P}). References: [1] \cite{li2018delete}, [2] \cite{suzgun2022prompt}.
The results of other systems are copied from prior study~\cite{suzgun2022prompt}.
}
\label{tab:amazon-clean}
\end{table}

\subsection{Effect of $\alpha$ in Prompt-then-AM}

\begin{figure*}[ht!]
    \centering
    \begin{minipage}{0.32\textwidth}
        \includegraphics[width=\linewidth]{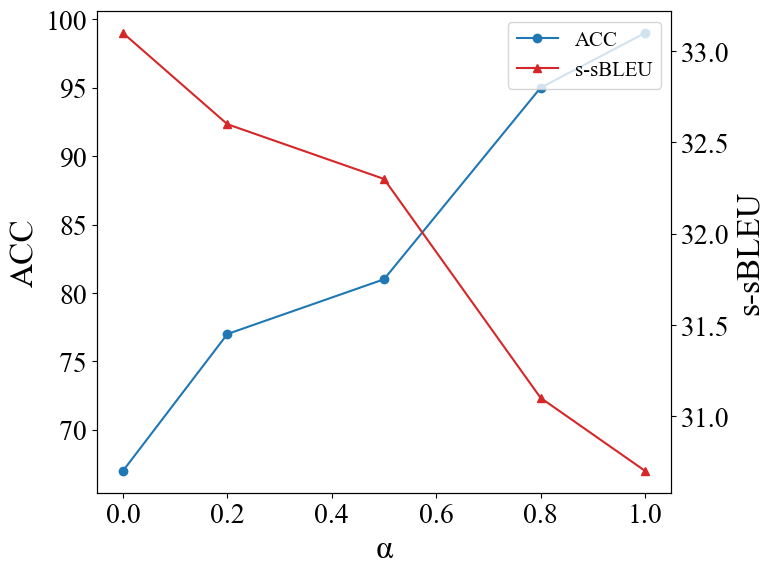}
        \caption{ACC and \textit{s}-sBLEU of Prompt-then-AM on Yelp-clean (\textit{N}$\rightarrow$ \textit{P}) with increasing $\alpha$.}
        \label{fig:acc_ssbleu}
    \end{minipage}\hfill
    \begin{minipage}{0.32\textwidth}
        \includegraphics[width=\linewidth]{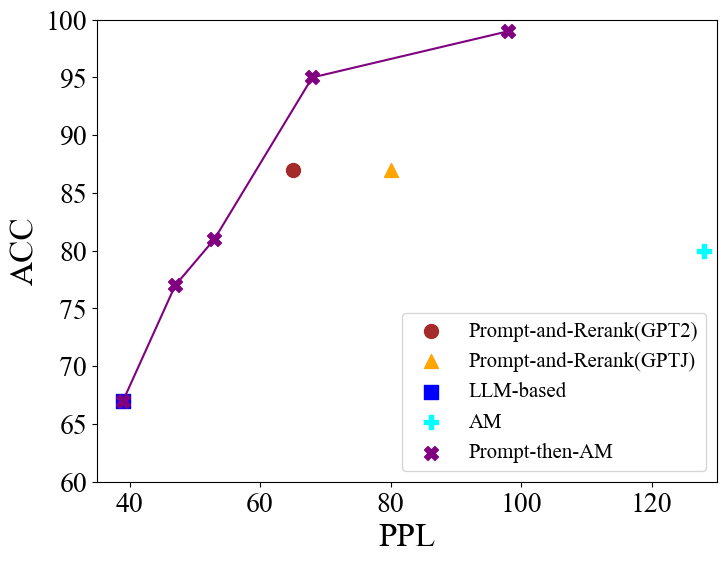}
        \caption{ACC and PPL of Prompt-then-AM on Yelp-clean (\textit{N}$\rightarrow$ \textit{P}) with various $\alpha$.}
        \label{fig:acc-ppl}
    \end{minipage}\hfill
    \begin{minipage}{0.32\textwidth}
        \centering
        \resizebox{\linewidth}{!}{%
        \begin{tabular}{l |c c c c }
        \toprule 
            Methods & \textbf{ACC} ($\uparrow$)  &	\textbf{\textit{r}-sBLEU} ($\uparrow$) &	\textbf{\textit{s}-sBLEU} ($\uparrow$) &	\textbf{PPL} ($\downarrow$) \\
        \midrule
        \multicolumn{5}{c}{ChatGLM2-6B} \\
        \midrule
            LLM-based & 67&	27.7& 33.1&	39\\
            Prompt-then-AM & 93&	26.7&	31.9&	59\\
            AM-then-prompt & $84$&	$30.1$ &	$36.4$ &	$53$\\
            \midrule
            \multicolumn{5}{c}{GPT2-XL} \\
            \midrule
            LLM-based & 22 & 12.7 & 15.1 & 83 \\
            Prompt-then-AM & 81 & 13.3 & 14.9 & 172 \\
            AM-then-prompt & 52 & 8.4 & 9.4 & 51 \\
            \midrule
            \multicolumn{5}{c}{GPT-J-6B} \\
            \midrule
            LLM-based & 46 & 12.8 & 28.9 & 49 \\
            Prompt-then-AM & 87 & 12.4 & 27.6 & 78 \\
            AM-then-prompt & 69 & 10.2 & 15.8 & 46 \\
            \bottomrule
        \end{tabular}%
        }
        \captionof{table}{A comparison of LLM-based methods with different backbone LLMs on Yelp-clean (\textit{N}$\rightarrow$ \textit{P}).}
        \label{tab:comparison}
    \end{minipage}
\end{figure*}



From the above experiments, we conclude that Prompt-then-AM shows a superior effect among the multiple interactions.
In Prompt-then-AM method, $\alpha$ serves as a crucial control parameters for modulating the extent of edit.
A higher $\alpha$ value signifies more aggressive edits while a lower value maintains the outputs of LLMs.
We draw the ACC and \textit{s}-sBLEU of Prompt-then-AM on Yelp-clean (\textit{N}$\rightarrow$ \textit{P}) with different $\alpha$ in Figure~\ref{fig:acc_ssbleu}.
When $\alpha$ equals to $0$, prompt-then-AM degrades into LLM-based method and when $\alpha$ equals to $1$, prompt-then-AM masks out all the intermediate outputs and re-prediction using AM model.
We observe an negative correlation between ACC and \textit{s}-sBLEU with the increasing $\alpha$.
This indicates that with the increasing number of edits on the outputs of LLMs, we can enhance the style strength of the transferred sentences at the expense of the changing the original semantics.

A positive correlation trend between ACC and PPL can be observed in Figure~\ref{fig:acc-ppl}.
With the increase of $\alpha$ value, ACC and PPL both increase, which indicates that the style strength is increased at the expense of decreasing the fluency.
We further compare Prompt-then-AM method with other UTST systems, a top left plot indicates a better results of high ACC and low PPL.
As we can see from the figure, Prompt-then-AM usually occupies the top left part, which indicates its capability of balancing the ACC and PPL metrics with varying $\alpha$.




\subsection{Different LLMs as Backbones}

We further try different LLMs as the backbone for the multi-ways interaction.
From Table~\ref{tab:comparison}, we observe the similar trends for various baselines.
Involving AM is able to improve the performance of LLM-based methods.
And Prompt-then-AM interaction is more efficient in generating comprehensively good outputs.
It is worth noting that different backbones may decide the upper bound of the UTST systems.
Methods with ChatGLM2 generally surpass the other systems, which indicates it is vital to select a good LLM as the collaborator for unsupervised text style transfer.



\subsection{Case Study}

\begin{table}[t!]
\centering
\resizebox{\linewidth}{!}{
\small
\begin{tabular}{l l}
\toprule 
Methods& Sentence\\
\midrule
Source text & there are no smiles and no customer \\
 &service.\\
\midrule
AM &there are {\color{red}always} smiles and great\\
& customer service.\\
LLM-based & there is {\color{red}a lack of} smiles and {\color{red}a lack of} \\
& customer service.\\
Prompt-then-AM & there are plenty of smiles and great\\
&  customer service.\\
AM-then-prompt & there is always a smile and {\color{red}exceptional}\\
& customer service.\\
 AM-as-demo& {\color{red}They make you feel at ease and their}\\
 & {\color{red}smiles} are contagious.\\
LLM-as-signal & {\color{red}delightful} smiles and customer service.\\
\midrule
Target text & there were plenty of smiles and cus-\\ 
& tomer service.\\
\bottomrule
\end{tabular}
}
\caption{Generated outputs of different methods on Yelp-clean (\textit{N}$\rightarrow$ \textit{P}).
The improper transferring is annotated with red color.
}
\label{tab:case_study}
\end{table}

In Table~\ref{tab:case_study}, we examine the outputs generated by different methods.
AM model is able to generate a sentence that is close to the target sentence.
LLM-based approach generates a sentence that conveys a slightly negative sentiment, which fails to produce a positive sentence.
AM-then-prompt method is able to partially change the sentiment but the unexpected revision by LLMs leads to a token ``\textit{exceptional}'', which cannot match the original semantic meaning.
LLM-as-signal largely changes the syntax of the sentence, leading to an undesirable prediction.
Unexpectedly, AM-as-demo dramatically change the source sentence.
This could happen when the demonstration for in-context learning misguides the semantic transfer to the LLMs. 
In comparison, Prompt-then-AM is more controllable in the perspective of preserving the semantics and transferring the polarity with rational edits.



\section{Conclusion}
\label{sec:conc}

In this paper, we investigate multi-way interaction of LLMs and attention masking for solving UTST tasks.
Specifically, we consider: pipeline framework with tuned orders; knowledge distillation from LLMs to attention masking model; in-context learning with constructed parallel examples.
We show UTST systems with combining LLMs and attention masking in four ways can improve the baselines on different evaluation metrics like style strength, content preservation, and text fluency.
We further show that simply conduct prompting and attention masking-based revision can consistently surpasses the other interactions and achieve SOTA results in Yelp-clean and Amazon-clean datasets even compared with supervised text style transfer systems.

\section*{Limitations}


LLMs have hallucinations. We have noticed that LLMs sometimes get confused by the instructions in a prompt and sometimes generate content that is completely unrelated to the input. They mix up the text that needs to be transformed with the instructions themselves, especially during In-Context Learning methods. This happens more when the prompts are too long and it is hard for the model to differentiate between instructions and the actual content.
In the future, we will dive into the research of easing hallucination issue in text style transfer and make more controllable text style transfer with LLMs.

\section*{Ethics Statement}

In our study on Unsupervised Text Style Transfer, no personal information was collected, ensuring the ethical integrity of our research. We emphasize that this endeavor involved no risk, as participants were neither exposed to harmful materials nor engaged in hazardous activities. Our commitment to ethical standards is unwavering.

\bibliography{latex/anthology,latex/custom}

\clearpage
\appendix
\section{Appendix}
\label{ap:implementation}

For training models, we use 2 NVIDIA V100 GPUs on a machine running Ubuntu 20.04 on a 10-Core Intel CPU. We choose RoBERTa-base as the baseline model for the classifier and trained on three datasets: Yelp, Amazon, and Politeness, respectively. For each training session, we set the batch size to 64, the number of epochs to 10, and the learning rate to 1e-5. We choose BART-base as our filling model and trained on different datasets. To prepare the training data for BART on the Yelp dataset, we set the mask predictor's $\alpha$ to 0.5. Similarly, for the Amazon dataset, we also set the $\alpha$ to 0.5. For the Politeness dataset, we set the mask predictor's $\alpha$ to 0.35. 
the detailed implementation inforamtion can be found in Table~\ref{tab:implementation}.

\begin{table}[h!]
\centering
\resizebox{\linewidth}{!}{
\small
\begin{tabular}{l|l l}
\toprule 
 Dataset & Parameters & Settings\\
\midrule
 RoBERTa-base&Batch size&64\\
 &Epochs& 20\\
 &Learning rate& 1e-5\\
 & $\alpha$&0.5 for Yelp and Amazon\\
 & &0.35 for Politeness\\
 \midrule
 BART-base&Batch size& 16\\
 &Epochs&  10 for Yelp and Amazon\\
 && 5 for Politeness\\
 \midrule
 ChatGLM2-6B&Temperature& 0\\
\bottomrule
\end{tabular}
}
\caption{Implementation details.
}
\label{tab:implementation}
\end{table}

\label{sec:appendix}

\end{document}